# Hand Hygiene Video Classification Based on Deep Learning

*Rashmi Bakshi*

*School of Electrical and Electronic Engineering*

*Technological University, Dublin, Ireland*

**Abstract**

In this work, an extensive review of literature in the field of gesture recognition is carried out along with the implementation of a simple classification system for hand hygiene stages based on deep learning solutions. A subset of robust dataset that consist of hand washing gestures with two hands as well as one-hand gestures such as linear hand movement is utilized. A pre trained neural network model, ResNet 50, with imagenet weights is used for the classification of 3 categories: "Linear hand movement", "Rub hands palm to palm " and " Rub hands with fingers Interlaced "movement. Correct predictions are made for the first two classes with > 60% accuracy. A complete dataset along with increased number of classes and training steps will be explored as a future work.

## 1    Introduction

Gesture interaction is very popular in the gaming industry with the use of gesture trackers and controllers such as Microsoft Kinect. However, besides the entertainment industry and other industries, such as manufacturing and automation, gesture interaction is expanding in the healthcare sector for assisting patients with mobility and other health-related issues [1]. One scenario where the tracking and identification of gestures are of interest is the process of hand washing. This has come to the fore in recent times because of the Covid-19 pandemic and the seasonal flu, where good hand hygiene practices are known to mitigate the spread of these viruses. The process of hand washing involves dynamic hand gestures. It may be possible to analyse these hand gestures and extract unique hand features for detection and classification with use of motion-based game controllers, gesture trackers, and cameras. The prospect of detecting the stages involved in the process of hand washing may be useful in a healthcare setting where it is proven that correct hand hygiene practices can reduce the rate of hospital-acquired infections [2]. In this work, deep learning solutions are adapted for the classification of various hand hygiene stages. A robust video-based hand washing dataset was created with the help of 30 participants in the past study. A small segment of this dataset is applied here.

## 2    Related work

Healthcare authorities such as the World Health Organization (WHO), the Centres for Disease Control and Prevention (CDC) have published structured guidelines for practicing hand hygiene in health care settings. Health care workers such as doctors, nurses and physicians should follow the standard approach for washing hands as per the advice of health care authorities [3].There are 11 sequential steps involving two hand gestures in the 'hand washing' guidelines as per WHO [4]. Hand gestures recognition data acquisition methods can be sub divided into vision based methods and device based methods. The previous work carried out for the implementation of real-time hand tracking is discussed below.

*Gesture recognition with 3D sensors*

Previous work with the use of various commercial gesture tracking devices such as Leap Motion Controller, Microsoft Kinect has been carried out for the purpose of detecting hand gestures in real-time. Fabio *et al*. [5] extracted the hand region from the depth map and segmented it into palm and finger samples. Distance features between the palm centre and the fingertips were calculated to recognize various counting gestures. Lin Shao [6] extracted the fingertip position and palm centre with the help of the Leap Motion Controller for tracking hand gestures. Distance between fingertip and palm centre and the distance between two fingers adjacent to each other was calculated. Velocity was detected to differentiate between static and dynamic gestures. In our previous work, Leap Motion Controller was utilised to differentiate between stationary and moving hand by extracting the palm velocity vector [7]. Marin *et al*. [8] used the Leap Motion Controller and the Microsoft Kinect jointly to extract the hand features such as fingertip position, hand centre, and hand curvature for recognizing the American Sign Language gestures. Jin *et al*. [9] used multiple Leap Motion sensors to develop a hand tracking system for object manipulation where the sum of distal interphalangeal, proximal interphalangeal and metacarpophalangeal angles were taken into account. Strength of grasping and pinching were the object manipulation tasks that were recorded. 3D gesture trackers have shown promising result in the past in context of gesture recognition. However, they were not explored for tracking hand hygiene stages. Rashmi *et al*. [8] has attempted to detect basic hand-hygiene stage, "Rub hands palm to palm" by extracting unique hand features such as palm orientation, palm curvature and distance between the two palms with threshold values.

*Gesture recognition with images*

The field of computer vision and image processing is greatly explored in the context of gesture recognition in the past. Vision based systems and applications were built for gesture tracking and detection. Khan [10] used colour segmentation and template matching technique to detect American Sign Language gestures. Jophin *et al*. [11] have developed a real time finger tracking application by identifying the red colour caps on the fingers using colour segmentation technique in image processing. Azad *et al*. [12] extracted the hand gesture by image segmentation and morphological operation for American Sign Language gestures. Cross –correlation coefficient was applied on the gesture to recognise it with overall 98.8 % accuracy. Chowdary *et al*. [13] detected the number of circles to determine the finger count in real-time, where in the scanning algorithm is independent of the size and rotation of the hand. Liorca *et al*. [14] has classified the hand hygiene poses using a traditional machine learning approach with a complex skin colour detection, particle-filtering model for hand tracking.

*Gesture recognition with deep learning*

Deep learning is an emerging approach and has been widely applied in traditional artificial intelligence domains such as semantic parsing, transfer learning, computer vision, natural language processing and more [15]. Over the years, deep learning has gained increasing attention due to the significant low cost of computing hardware and access to high processing power (eg-GPU units) [15]. Conventional machine learning techniques were limited in their ability to process data in its natural form. For decades, constructing a machine learning system required domain expertise and fine engineering skills to design a feature extractor that can transform the raw data (example: pixel values of an image) into a feature vector, which is passed to a classifier for pattern recognition [16]. Deep learning models learn features directly from the data without the need for building a feature extractor. Researchers have utilised deep learning/transfer learning for monitoring hand gestures. Yeung *et al.* [17] presented a vision-based system for hand hygiene monitoring based on deep learning. The system uses depth sensor instead of the full video data to preserve privacy. The data is classified using a convolutional neural network (CNN). Li *et al*. [18] conducted an investigation of CNN for gesture recognition and achieved high accuracy, showing that this approach is suitable for the task. Yamamoto *et al.* [19] use vision-based systems and a CNN for a hand-wash quality estimation. They compare the quality score of the automated system with a ground-truth data obtained using a substance fluorescent under ultraviolet light. Their results show that the CNN is able to classify the washing quality with high accuracy. Ivanovs *et al.* [20] train the neural network on labelled hand washing dataset captured in a health care setting , apply pre trained neural network models such as MobileNetV2 and Xception with  >64% accuracy in recognising hand washing gestures.

## 3     Dataset

A robust dataset that consists of various hand hygiene gestures was recorded with the help of 30 participants. Each participant was informed about the hand hygiene steps as per WHO and videos were recorded of their hand movements while they imitated the steps in a laboratory sink in the presence of a soap dispenser and a water tap [21]. In addition to six hand washing gestures, one hand gestures such has linear hand movement and circular hand movement were recorded. In this paper, a small subset of a dataset is used to test the accuracy of deep learning models in classification and prediction of hand hygiene videos. In future, the full potential of deep learning/ transfer learning will be explored by utilising the complete dataset. Hand hygiene videos were segmented into smaller video set with individual hand hygiene gesture. The videos were further converted into frames/ images with the help of Python and OpenCV. Three classes were prepared, linear movement, Fingers interlaced and palm-to-palm with 162 jpeg files.

## 4     Transfer learning

  Deep learning models essentially require thousands of data samples and heavy computational resources such as GPU for accurate classification and prediction analysis. However, there is a branch of machine learning, popularly known as "transfer learning" that does not necessarily requires large amounts of data for evaluation. Transfer learning is a machine learning technique wherein a model developed for one task is reused for the second related task. It refers to the situation where "finding" of one setting is exploited to improve the optimisation in another setting [22].Transfer Learning is usually applied to the new dataset, which is generally smaller than the original data set used to train the pre-trained model. Hussain *et al*. [22] applied transfer learning to train caltech face data set with 450 face images with pre-trained model on ImageNet data set. Increasing the number of training steps (epochs) increased the classification accuracy but increased the training time as well. Computational power and time were the main limitations within the study.

Keras API [23] provides the most common workflow of transfer learning in context of deep learning. They are:

1. Take layers from a previously trained model.
2. Freeze them to avoid destroying any of the information that they contain during the future training rounds.

3. Add some new, trainable layers on top of the frozen layers. They will learn to turn the old features into predictions on a new dataset.
4. Train the new layers on your dataset.

Based on this workflow, a pre-trained CNN model, ResNet50 [50 layers deep] on ImageNet weights is applied in this work. The network is pre- trained on more than one million images of ImageNet dataset and can classify images into 1000 object categories. `Weights="imagenet"; include_top= False` was selected as the head of the model was replaced with a new set of fully connected layers with random initializations. All layers below the head are frozen so that their weights cannot be updated. `layer.trainable = False`. The model implementation is adapted from [24] where the author has implemented a classification system for 'sports' related video recordings.

## 5    Results and Discussion

The network is trained for 50 training steps (epochs) meaning that the network is fed with the data in a loop of 50 cycles in order to establish and learn the underlying patterns and features. As ResNet is a deep network, the amount of time taken for the training was > 2 hours even for the smallest dataset used in this study. This experiment serves as a preliminary case study to demonstrate the potential of deep learning solutions for a complex problem of hand hygiene video classification that comprise of six dynamic two hand gestures. The classification report is presented in Table 5.1. It clearly indicates that the network is trained correctly for classes Linear and Finger Interlaced with a precision, recall and f1 score >0.5. The network scores poorly for 'Rub hands palm to palm' hand movement and it is evident from the class predictions in Figure 5.4. Figure 5.1 is the plot for loss-accuracy curve for training and validation dataset obtained after the model is trained and saved. It shows that the loss curve decreases from the range of 1.2-1.4 in 1$^{st}$ epoch to the range, 0.6-0.9 by 50$^{th}$ epoch (training step) for both sets. However, the loss value is still too high > 0.5. In an ideal case, as training steps increase, the loss function decreases to 0 and the accuracy function maximises to 1. Figure 5.2 and figure 5.3 is the representation of the correct predictions for the linear hand movement and Rub Palms with fingers interlaced. For future, this work will be expanded to incorporate all the hand hygiene stages with six categories with the use of complete hand washing dataset.

|  | Precision | Recall | F1 score | Support |
|---|---|---|---|---|
| **FingersInterlaced** | 0.54 | 1.00 | 0.70 | 14 |
| **Linear** | 0.93 | 1.00 | 0.97 | 14 |
| **Palm2Palm** | 0.00 | 0.00 | 0.00 | 13 |
|  |  |  |  |  |
| **Micro avg** | 0.68 | 0.68 | 0.68 | 41 |
| **Macro avg** | 0.49 | 0.67 | 0.56 | 41 |
| **Weighted avg** | 0.50 | 0.68 | 0.57 | 41 |

Table 5.1 Classification report for classes- Rub palms with Fingers Interlaced, Linear hand movement and Rub hands palm to palm.

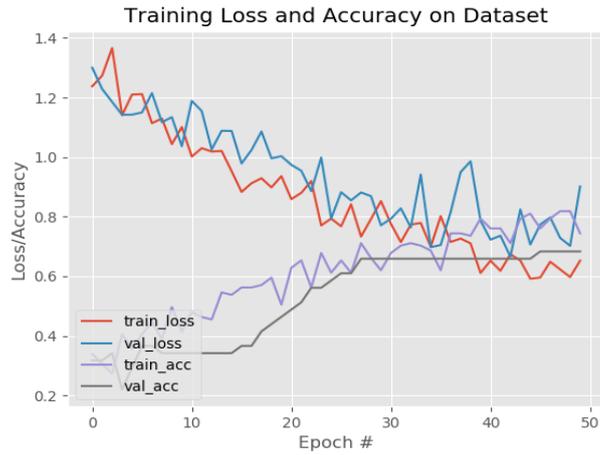

Figure 5.1: The loss/ accuracy curve with >60% accuracy

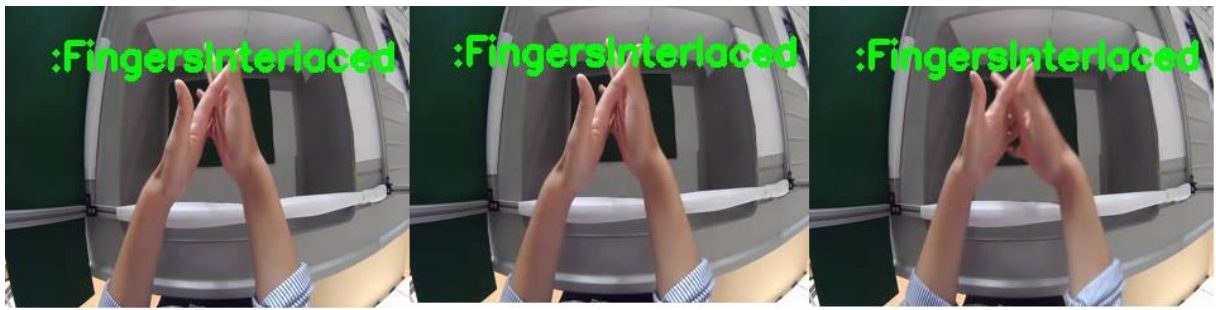

Figure 5.2: Correct prediction for the selected frame no 38, 60, 64- FingerInterlaced.avi

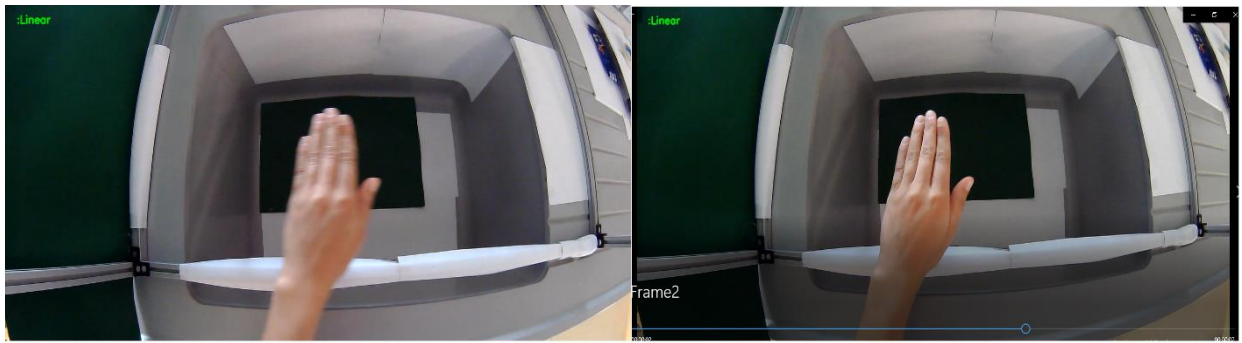

Figure 5.3: Correct prediction for "Linear.avi" frames.

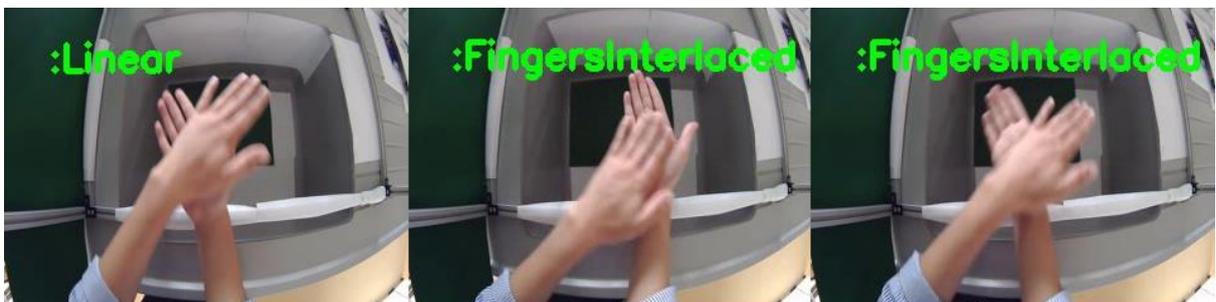

Figure 5.4: Incorrect prediction for Rub palm to palm frame no. 45, 55, 60 shown- palmtopalm.avi